# PCD-ReID: Occluded Person Re-Identification for Base Station Inspection

Ge Gao
Petroleum Institute
China University of Petroleum
(Beijing) at Karamay
Karamay,China
2609565842@qq.com

Zishuo Gao
Engineering Institute
China University of Petroleum
(Beijing) at Karamay
Karamay, China
2023016002@st.cupk.edu.cn

Hongyan Cui
School of Information and
Communication Engineering
Beijing University of Posts and
Telecommunications
Beijing, China
cuihy@bupt.edu.cn

Zhiyang Jia
Petroleum Institute
China University of Petroleum
(Beijing) at Karamay
Karamay, China
2018592015@cupk.edu.cn

Zhuang Luo
Petroleum Institute
China University of Petroleum
(Beijing) at Karamay
Karamay,China
2024216807@st.cupk.edu.cn

ChaoPeng Liu
Petroleum Institute
China University of Petroleum
(Beijing) at Karamay
Karamay,China
2024216805@st.cupk.edu.cn

*Abstract*—Occluded pedestrian re-identification (ReID) in base station environments is a critical task in computer vision, particularly for surveillance and security applications. This task faces numerous challenges, as occlusions often obscure key body features, increasing the complexity of identification. Traditional ResNet-based ReID algorithms often fail to address occlusions effectively, necessitating new ReID methods. We propose the PCD-ReID (Pedestrian Component Discrepancy) algorithm to address these issues. The contributions of this work are as follows: To tackle the occlusion problem, we design a Transformer-based PCD network capable of extracting shared component features, such as helmets and uniforms. To mitigate overfitting on public datasets, we collected new real-world patrol surveillance images for model training, covering six months, 10,000 individuals, and over 50,000 images. Comparative experiments with existing ReID algorithms demonstrate that our model achieves a mean Average Precision (mAP) of 79.0% and a Rank-1 accuracy of 82.7%, marking a 15.9% Rank-1 improvement over ResNet50-based methods. Experimental evaluations indicate that PCD-ReID effectively achieves occlusion-aware ReID performance for personnel in tower inspection scenarios, highlighting its potential for practical deployment in surveillance and security applications.

*Keywords—person re-identification, component discrepancy, feature extraction, deep learning*

## I. INTRODUCTION

The task of re-identifying inspection personnel in a base station monitoring environment involves using advanced computer vision techniques to swiftly locate images of a specific inspection worker from a large database, matching their identity accurately [1]. This task falls under the domain of computer vision, specifically as a subtask of image retrieval. Recently, with increasing demand in industrial safety, intelligent surveillance, and criminal investigations, this topic has gained significant research interest [2],[3]. For this study to be practical in seamless workplace attendance systems, two critical factors determine its feasibility: accuracy and time tolerance. Regarding accuracy, the base station surveillance system images, as shown in Fig. 1, are affected by various challenges, including illumination changes, viewpoint differences, complicated backgrounds, and occlusions, which introduce instability to pedestrian re-identification algorithms. In such complex environments, traditional re-identification methods often fail to capture the dynamic information such as illumination changes, viewpoint differences, and complicated backgrounds, and lack the ability to focus on shared visible features of occluded individuals. Consequently, these methods typically fall short of satisfactory performance, as they cannot effectively extract common, unoccluded features [4]. Therefore, the study of occlusion-aware re-identification for inspection personnel in such environments presents a highly valuable research avenue.

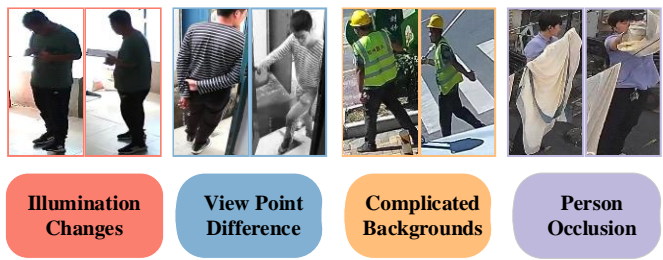

Fig. 1. The re-identification of inspection personnel in base station environments encounters several key challenges. First, significant illumination changes arise due to the substantial difference in lighting conditions inside and outside the base station. Second, viewpoint variations are introduced by the limited number of isolated cameras inside the station. Third, complex background issues emerge from the more diverse panoramic views captured by cameras positioned outside the

This work was supported by National Natural Science Foundation of China (62171049), China University of Petroleum (Beijing) Karamay Campus introduction of talents and launch of scientific research projects (XQZX20240010), and China Tower Corporation Limited IT System 2023 Package Software Project - AI Algorithm and Services.

XXX-X-XXXX-XXXX-X/XX/$XX.00 ©20XX IEEE

base station. Finally, partial occlusions, which are often unavoidable, add to the difficulty of the re-identification process.

In the field of pedestrian re-identification (ReID) for specific domains, many researchers have explored various methods to work with existing datasets [5],[6],[7]. Earlier studies attempted to generalize models trained on one domain to apply in a specific target domain [8],[9], This approach has proven effective and innovative [10],[11], significantly improving model generalization, though accuracy still requires enhancement. Researchers have also observed that models trained on large-scale datasets, such as MSMT17[12], typically perform better when transferred to smaller-scale datasets, like Market1501, than the reverse[13],[14],[15],[16]. Overall, for ReID to function as a seamless attendance solution in base station environments, a practical approach would be to train a large-scale ReID model on an extensive dataset. Such a model, fed with vast amounts of data, could achieve robust performance even in previously unseen base station environments. However, this solution presents three critical challenges: collecting a sufficiently large dataset, designing a deep network to learn from it, and evaluating the model's real-world performance.

To address the challenges of occluded ReID in base station environments, this paper proposes the PCD-ReID algorithm, with the following main contributions: We collected real-world patrol surveillance images for model training over six months, capturing more than 10,000 individuals and 50,000 images. The raw data was manually annotated and formatted similarly to the Market1501 dataset [17] to accommodate various model training processes. Additionally, we designed a Pedestrian Component Discrepancy network based on the Fast-ReID framework [18], employing multiple Vision Transformer structures to extract pedestrian features in base station environments, and supervised the training process using multiple loss functions. Our approach diverges from domain generalization; instead, it introduces a vertical industrial model specifically tailored for high-performance, lightweight deployment. Furthermore, simulations demonstrate that our model achieves a mean Average Precision (mAP) of 79.0% and a Rank-1 accuracy of 82.7% on an occlusion-based base station ReID test dataset. Experimental results verify that PCD-ReID is highly effective for handling occlusion, low resolution, complex angles, and lighting variations within base station environments, and the algorithm has already been practically implemented. The subsequent sections of this paper will discuss these contributions in further detail.

## II. RELATED WORK

Due to the significance of occluded pedestrian re-identification, numerous researchers have made valuable contributions and proposed various methods. In this section, we present key works relevant to our study, including occluded pedestrian re-identification algorithms, Vision Transformer models, and the Triplet Attention Module, based on a personal perspective.

### A. Occluded Person ReID Algorithm

The primary goal of occluded pedestrian re-identification algorithms is to improve identification accuracy in environments with occlusions [19]. With advancements in deep learning and GPU computing power, numerous effective architectures have emerged in the field of pedestrian re-identification. These architectures typically rely on a backbone network with additional branch structures to extract local features [20] . Backbone networks for global feature extraction generally fall into two categories. The first category includes convolutional neural network (CNN)-based models [21],[22], such as ResNet, which performs well in non-occluded environments [19],[23]. These CNN-based models typically excel at extracting global features [24], but they struggle to handle occlusions effectively, as they may mistakenly incorporate occlusions as part of the main subject, resulting in erroneous features [25],[26],[27]. The second category involves models based on attention mechanisms, which are generally more robust for occluded pedestrian re-identification [28],[29],[30]. These Transformer-based models can better focus on unobstructed shared parts, enhancing the extraction of partial features [31].

### B. Vision Transformer Model

Occluded pedestrian re-identification in complex environments presents a significant challenge. Traditional ReID methods typically rely on global feature extraction, leading to substantial performance degradation when partial occlusions appear in pedestrian images. To address this issue, Vision Transformer (ViT)-based models have emerged as effective solutions in recent years [1],[32],[33]. ViT, with its powerful global modeling capabilities, captures long-range dependencies within images, enhancing the recognition of occluded pedestrians.

In the field of occluded ReID, substantial research has focused on designing robust feature extraction methods to address partial occlusions. Some early studies employed local feature learning approaches, extracting features based on different body parts to avoid the loss of global features due to occlusions [34]. However, these methods depend on precise part detection, and detection inaccuracies can negatively impact identification outcomes. In contrast, ViT-based ReID methods use self-attention mechanisms to adaptively focus on various image regions [35], eliminating the need for precise part detection and more effectively addressing occlusions.

Recently, ViT-based methods, such as TransReID [36], have achieved significant performance improvements across multiple datasets. TransReID enhances the standard ViT by incorporating image partitioning and spatial position encoding, bolstering the model's robustness against partial occlusions. Additionally, the strong performance of Transformer-based ReID models in handling occlusion scenarios has spurred research into further optimization and broader application of these methods.

In summary, ViT-based ReID methods demonstrate great potential in occluded environments. Future research may explore the integration of multimodal data and external semantic information to improve robustness in complex occlusion scenarios.First, confirm that you have the correct template for your paper size. This template has been tailored for output on the US-letter paper size. If you are using A4-sized paper, please close this file and download the file "MSW_A4_format".

## C. Triplet Attention Module

The pedestrian re identification task in the small room scene of base stations often encounters challenges such as viewpoint variations, lighting conditions, and occlusions. The Triplet Attention Module [37] addresses these issues by capturing multi-dimensional interactive information and integrating features with Transformers, enhancing model robustness and representation capacity for improved recognition accuracy. Within Transformer-based models, the Triplet Attention Module strengthens information fusion across different feature maps. By capturing interactions between channel and spatial dimensions [38], triplet attention helps the model better understand both global and local pedestrian features [39], which is critical for ReID. Transformer models rely on self-attention mechanisms to capture relationships between different parts of the input sequence. The Triplet Attention Module can further enhance this representational ability, especially when processing image data, as it enables the model to learn richer spatial and channel relationships, improving the discriminative power of pedestrian features [40]. In this study, we employ a triplet loss function during training to further optimize model performance. The triplet loss function aids in learning more distinctive feature representations by pulling closer the feature vectors of the same identity while pushing apart those of different identities.

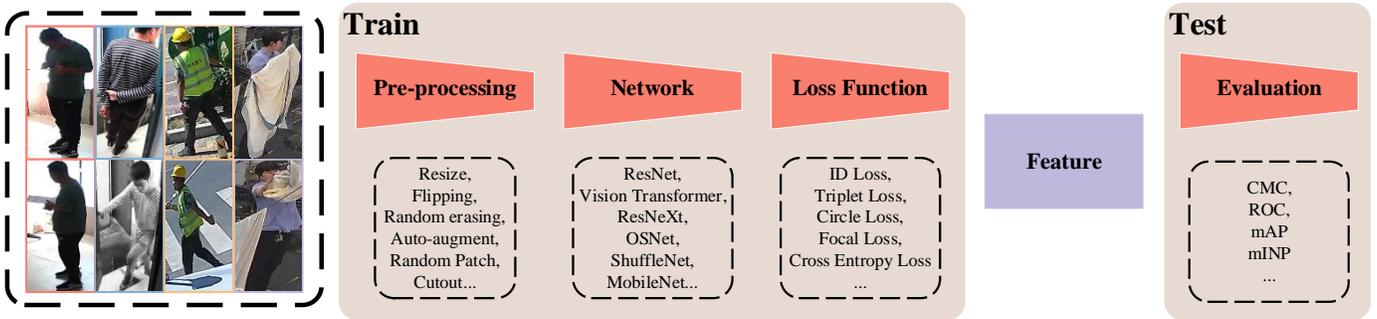

Fig. 1. The framework used in this study to address occluded re-identification of base station inspection personnel is shown in the diagram above. After inputting data, the process begins with the training phase, during which the model is trained to compute pedestrian image feature values. In the testing phase, the model's performance is evaluated. In the training phase, data preprocessing is first conducted, followed by input into various network structures, and supervised learning is achieved through multiple loss functions. Data preprocessing includes basic image operations such as resizing, flipping, random erasing, augmentation, patching, and cutout. For network structures, we can compare several backbone networks commonly used in pedestrian re-identification, including ResNet, Vision Transformer, ResNeXt, OSNet, ShuffleNet, and MobileNet. The model's parameter adjustments are supervised by multiple loss functions, such as ID Loss, Triplet Loss, Circle Loss, Focal Loss, and Cross Entropy Loss. In the testing phase, the model extracts feature values from images and performs distance matching within the gallery, with distance ranking to evaluate model effectiveness using metrics such as CMC, ROC, mAP, and mINP.

## III. PEDESTRIAN COMPONENT DIFFERENCES REID

This section provides a detailed explanation of our proposed PCD-ReID algorithm process, as illustrated in Fig. 2, covering data preprocessing, the training phase, and the testing phase. Our proposed PCDNet incorporates a Triplet Attention mechanism at the forefront of ViT, applied before patching. The Triplet Attention module processes the input data without altering its dimensions. After patching, the ViT model, consisting of 12 blocks, outputs a 768-dimensional feature vector for each image. These model parameters are fine-tuned under the guidance of combined loss functions. Our algorithm demonstrates strong performance in optimizing the model during the training process.

### A. Preprocessing For MyTT2 Dataset

To better align our model with real-world application scenarios, we collected a new dataset of real-world inspection surveillance images over a span of six months. This dataset includes images from different inspection personnel, with over 10,000 identities and more than 50,000 images, averaging around six images per person. A primary focus during data collection was to capture occluded images, and this dataset is named MyTT2. To ensure compatibility of MyTT2 with other models, we adapted its format to align with the commonly used Market1501 dataset [41],[42]. This compatibility enables seamless integration of Market1501 data for training on MyTT2, significantly saving time for model training and evaluation.

The structure of the MyTT2 dataset, illustrated below, includes three main subsets: bounding_box_test, bounding_box_train, and query. bounding_box_test serves as a gallery image pool for retrieving images from the query set. bounding_box_train consists of training images from around 10,000 different identities, while query is a set of probe images, each representing a unique identity captured by different cameras.

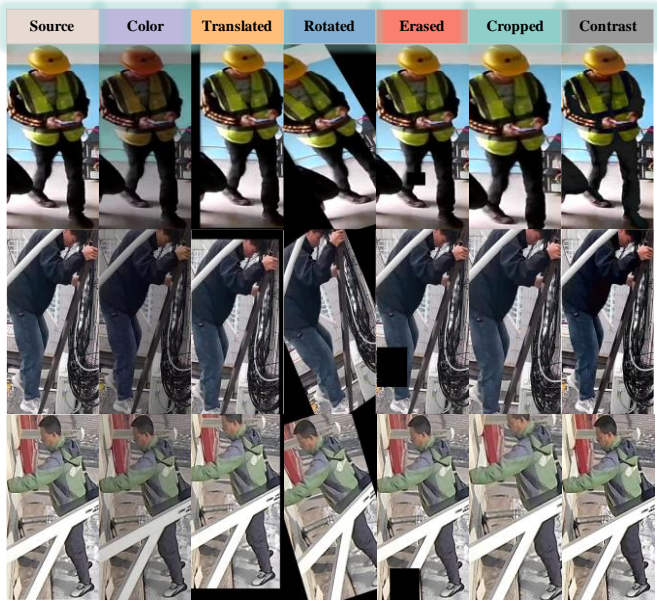

Fig. 2. Illustration of data augmentation operations. To enable the model to learn more effectively in the occluded and complex environment of a base station room, we implemented extensive data augmentation techniques. From left to right in the Fig. 3, the operations are as follows: source data; color jitter to help the model adapt to color variations due to lighting changes; image translation to simulate scenarios where the subject might move out of the camera frame; rotation to broaden the dataset with varied subject orientations; random erasing to mimic cases where the subject may be partially occluded; scaling to address variations in the inspector's distance from the camera; and contrast adjustment to enhance model robustness to lighting changes. These augmentation operations significantly increase both the quality and quantity of training data.

To expand the dataset further, we implemented automatic data augmentation, random data augmentation, and enhanced augmentation, as shown in Fig. 3. These augmentations include a wide range of image transformations commonly used in image classification, providing varied strategies and parameter configurations. Specific operations include rotation, cropping, translation, color adjustment, and contrast adjustment. These transformations do not alter image dimensions, allowing the augmented images to be directly input into the model for training. This approach significantly enhances the model's robustness and generalization. The code also includes default augmentation strategies, parameter configurations, and functions for generating random augmentations.

## B. PCD-Network

The PCDNet feature extraction network presented in this paper is a fine-tuned version of the Vision Transformer (ViT) architecture, enhanced with specialized modules such as the Triplet Attention module to improve its effectiveness. ViT is a neural network architecture based on self-attention, which segments images into a series of patches and uses a Transformer model to learn image representations. The Fig. 4 below provides a detailed view of the structure of our PCDNet.

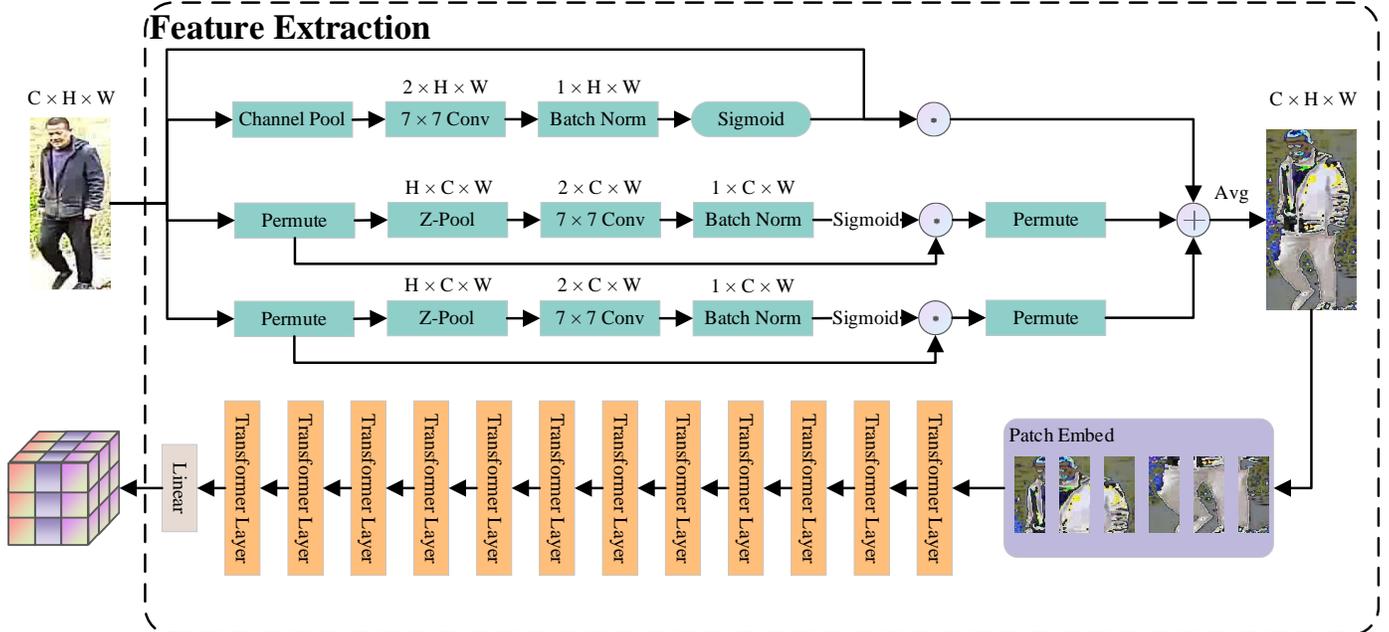

Fig. 4. Our proposed PCDNet processes input images sequentially through a Triplet Attention module and a Vision Transformer module. The Triplet Attention module consists of three branches, each capturing semantic information from different perspectives: Channel, Height, and Width. After obtaining these three sets of semantic information, attention from each direction is fused through weighted combination. As shown in the diagram, the Triplet Attention module effectively identifies the edges and contours of the person and their components. Next, the Patch Embed module transforms the high-dimensional image information into lower-dimensional semantic information, which passes through 12 Transformer layers before a final classification layer extracts the image feature information.

In PCDNet, three highly effective structures are employed: the Triplet Attention Mechanism, a Patch Embedding structure, and multiple Transformer Layers. The first component, the Triplet Attention Mechanism, shown in light blue, includes the Permute operation, which rearranges the dimensional order of the data. Batch Normalization is used to accelerate training and enhance model stability, while the Sigmoid activation function scales the output between 0 and 1. An Average Pooling layer reduces the spatial dimensions of the feature map, and Channel Pooling fuses features along the channel dimension.

The Patch Embed layer segments the input image into small patches and embeds these patches into a high-dimensional space. The Transformer layers, based on a self-attention mechanism, capture long-range dependencies within the data. Additionally, Z-Pool offers two main advantages: dual statistics merging and channel dimension fusion. Z-Pool calculates both the maximum and average values of the input tensor and then merges these statistics along the channel dimension. This approach captures both global and local features, enhancing the model's ability to represent features at various scales. By merging maximum and average values along the channel dimension, the model can increase the number of channels without changing the spatial dimensions of the feature map, facilitating further processing and integration of these statistical features in subsequent network layers.

In PCDNet, combining these three structures provides an efficient approach to handling image data, enhancing model performance and generalization through the fusion of local feature extraction and global context understanding.

*C. Joint Loss Function*

To address the challenge of rapid convergence with hard triplet loss, which can lead the model to prematurely stop learning new information, we propose a complementary, robust, and flexible loss function named $L_{\text{pcd}}$. This loss function combines the strengths of hard triplet loss [43], cross-entropy loss, and circle loss [44], providing comprehensive supervision for the model. The circle loss component enhances intra-class compactness and inter-class separability, contributing to notable accuracy improvements. Each of these loss functions targets different aspects of model performance: for example, the triplet loss focuses on bringing positive samples closer while pushing negative samples further, and the cross-entropy loss emphasizes category probability accuracy. Together, they ensure multi-faceted optimization of the model. Additionally, we explore the effect of Smoothed Cross-Entropy loss, which also contributes substantially to improved performance.

Our proposed joint loss function for occluded re-identification of base station inspection personnel is expressed as:

$$L_{pcd} = \alpha L_{ce} + \beta L_{tri} + \gamma L_{cir} + \delta L_{cos} \quad (1)$$

where $L_{ce}$ is defined as:

$$L_{ce} = -\sum_{i=1}^{N} y_i \log \hat{y}_i \quad (2)$$

with $y_i$ representing the true label and $\hat{y}_i$ the predicted probability. This loss is widely used in classification tasks and measures the divergence between predicted and true category distributions, aiding in the supervision of model classification performance.

The triplet loss $L_{tri}$ is formulated as:

$$L_{\text{tri}} = \sum_{i=1}^{N} \left[ |f(x_i^a) - f(x_i^p)|^2 - |f(x_i^a) - f(x_i^n)|^2 + \alpha \right]_{+} \quad (3)$$

where $f(x_i^a)$ is the anchor sample, $f(x_i^p)$ the positive sample, $f(x_i^n)$ the negative sample, and α margin parameter. The triplet loss is instrumental in learning discriminative features by reducing the distance between positive samples and increasing it between negative samples, especially useful for tasks like face recognition and image retrieval.

The circle loss $L_{cir}$ is defined as:

$$L_{\text{cir}} = \log\left[1 + \sum_{(i,j)\in\mathcal{P}} e^{\gamma(s_{i,j} - \Delta_p)} + \sum_{(i,k)\in\mathcal{N}} e^{\gamma(\Delta_n - s_{i,k})}\right] \quad (4)$$

where γ is a scaling factor, $s_{i,j}$ the positive sample score, $s_{i,k}$ the negative sample score, and $\Delta_p$ and $\Delta_n$ the positive and negative margins, respectively. Circle loss excels in similarity learning by promoting tight clustering of similar samples and greater distinction of dissimilar ones, beneficial in handling hard samples in both classification and retrieval tasks.

Lastly, Cosine Loss $L_{cos}$ is defined as:

$$L_{\cos} = -\frac{1}{N}\sum_{i=1}^{N} \log \frac{e^{s(\cos\theta_{y_i} - m)}}{e^{s(\cos\theta_{y_i} - m)} + \sum_{j \neq y_i} e^{s\cos\theta_j}} \quad (5)$$

where $\theta_{y_i}$ is the predicted angle, m is the margin, and s the scaling factor. Cosface Loss is effective for enhancing inter-class separation by projecting samples onto a hypersphere, narrowing intra-class differences and expanding inter-class separation, especially suitable for high-dimensional classification tasks like face recognition.

By combining these loss functions, the model achieves balanced classification and feature discrimination, ultimately enhancing performance in complex tasks.

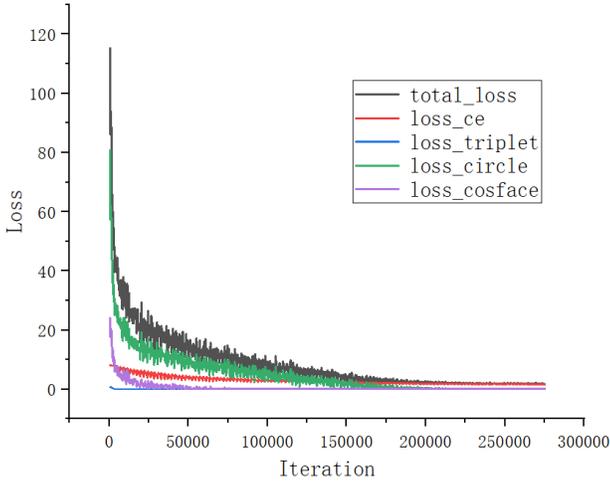

Fig. 5: The use of a combined loss function offers greater advantages over a single loss function. During convergence, the cross-entropy loss function reaches a certain point where it no longer decreases, while the triplet loss function starts with a relatively low initial loss value, limiting further reduction. However, by incorporating the circle loss and Cosface loss, we avoid issues where the initial loss is low and remains stagnant. This approach provides more effective feedback for adjusting model parameters.

The Fig. 5 demonstrates the superiority of the circle loss function over the hard triplet loss function in base station pedestrian re-identification. Using multiple loss functions can significantly enhance the model's generalization ability. Some loss functions may be more sensitive to outliers or difficult samples, while others are more robust; combining them helps reduce the risk of overfitting to specific types of data. Each loss function can be adjusted according to different data distributions and task requirements. For example, the circle loss function improves feature discrimination by optimizing intra- and inter-class similarities, whereas the Cosface loss function enhances model discrimination by introducing a cosine margin. In some cases, a single loss function may not sufficiently capture the complexity of the data, and combining multiple loss functions enables more comprehensive feature extraction, thereby improving overall model performance. This method was validated through simulation experiments on a large-scale base station dataset, achieving promising results on the occlusion dataset, with Rank-1 accuracy reaching 82.7% and mAP at 79.0%.

Addressing an engineering-related issue observed during training, both Circle Loss and Cosface Loss remained at zero from the start until the end of training, never exceeding zero. However, an interesting finding emerged: setting the "NUM of INSTANCE" parameter to a value greater than 1 immediately resolved the issue, allowing both Circle Loss and Cosface Loss to obtain normal values. For further analysis, reducing sample diversity was examined in detail. The "NUM of INSTANCE" parameter specifies the number of times each identity appears in each batch. When "NUM of INSTANCE" is set to 1, each identity appears only once per batch, reducing sample diversity and potentially limiting the model's ability to learn distinctive identity features, impacting loss function calculation. Setting "NUM of INSTANCE" to 2 means each identity has two samples per batch, enabling the model to learn more robust features by processing multiple views of each identity in each batch.

The calculation of loss functions is sensitive to sample quantity within batches. When the sample count decreases, loss value stability may suffer, resulting in anomalous calculation outcomes. Reduced sample counts can affect loss function gradient computation, especially for pair-based losses like Circle Loss and Cosface Loss, as fewer samples may reduce the number of effective sample pairs, impacting the calculated loss values.

IV. EXPERIMENTS

A. Experimental Setups

The software and hardware configuration for this experiment is as follows: Lenovo Legion Y7000P 2023 model, CPU: i7-13620H, GPU: RTX4060, RAM: 40GB, Storage: 1TB SSD. The software environment includes PyCharm 2023.2.1, Python 3.8.18, and PyTorch 1.7.1.

In addition to the hardware setup, the experimental details include the following: all images in the training and testing sets were resized to 256×128, with data augmentation applied at a probability of 0.1. The default dataset used is MyTT2. The optimizer for PCDNet is SGD, and each network was trained for 120 epochs with an initial learning rate of 0.008, utilizing cosine learning rate decay. We evaluated our model's performance on the MyTT2 dataset.

B. Comparative experiment

To provide reliable data supporting the performance benefits of the proposed algorithm, this paper conducts comparative experiments to highlight the capabilities of PCDNet. First, a baseline backbone network was chosen from classic networks, such as ResNet and OSNet, which are widely regarded as effective in prior research. To ensure an objective evaluation of PCDNet's effectiveness, we utilized the Cumulative Matching Characteristic (CMC) curve and mean Average Precision (mAP) as performance metrics, with the CMC curve specifically assessing Rank-1, Rank-5, and Rank-10. Detailed descriptions of these metrics are provided below.

The CMC curve measures the accuracy of retrieving results of the same identity within the top $n$ ranked results, which corresponds to the hit probability for Rank-$k$. Rank-$k$ refers to the probability that a correct match appears in the top $k$ search results (with highest confidence). It is calculated as follows:

$$Rank(k) = \frac{\sum_{probe \in Q} f(probe, k)}{N_q} \quad (6)$$

where $f(probe, k)$ returns 1 if there is a correct match among the top k results returned by the retrieval system, and it returns 0 if there is no correct match within those top k results.

Here, *probe* is the person image query from the query set $Q$, and $N_q$ is the number of queries. It is worth noting that Rank-$k$ indicates only the probability of a correct sample appearing in the top $k$ retrieval results and does not reflect the quantity of correct samples or the stability of the model, so an additional metric, mAP, is required for evaluation.

Mean Average Precision (mAP) is commonly used to assess a model's accuracy and robustness. Specifically, for a given probe image, similarity is computed between it and images in the candidate gallery, then ranked in descending order. If the $k$-th image in this ranking has the same ID as the probe, the precision $P_k, probe$ for the top $k$ returned results is calculated as follows:

$$P(k, probe) = \frac{k_c}{k} \quad (7)$$

where $k_c$ is the number of true positives in the top $k$ images. Based on this precision, the average precision $AP_{probe}$ for the probe is computed as follows:

$$AP(probe) = \frac{\Sigma_{j \in j_1, j_2, \dots, j_M} P(j, probe)}{M} \quad (8)$$

where $M$ is the number of true positive images in the gallery for the probe, and $\{j_1, j_2, \dots, j_M\}$ are the indices of these true positives within the ranking.

TABLE I
EXPLANATION OF OUR ALGORITHM'S SUPERIORITY ON MYTT DATASET

| Method | Rank-1 | Rank-5 | Rank-10 | mAP |
|---|---|---|---|---|
| ResNet50 | 66.8 | 81.6 | 85.4 | 61.7 |
| ResNsSt50 | 62.8 | 76.7 | 81.4 | 57.4 |
| OSNet [45] | 67.0 | 78.4 | 81.9 | 58.1 |
| ViT[36] | 69.5 | 84.2 | 87.8 | 53.9 |
| VersReID[1] | 66.3 | 73.4 | 82.7 | 85.3 |
| PGFA[26] | 52.8 | 63.8 | 67.8 | 42.1 |
| HOReID[19] | 54.1 | 61.5 | 72.1 | 42.7 |
| RFCnet[22] | 58.5 | 68.6 | 68.2 | 41.5 |
| PCDNet (Ours) | **82.7** | **92.2** | **94.5** | **79.0** |

As shown in Table I, PCDNet achieves a Rank-1 accuracy of 82.7%, with Rank-5 and Rank-10 reaching 92.2% and 94.5%, respectively, and an mAP of 79.0%. These results outperform other comparison methods, underscoring the effectiveness and superiority of PCDNet for the challenging task of re-identifying occluded inspection personnel at base stations. In contrast to a standard ViT, PCDNet incorporates a triplet attention mechanism before the patch embedding step. This mechanism captures semantic information from three perspectives—Channel, Height, and Width—and applies weighted fusion of attention in each direction. This design enables the model to better capture edges and contours of human features, thus enhancing feature extraction capabilities.

V. ALGORITHM EXPLAINABILITY

*A. Ablation Study*

This study integrates various beneficial modules into the model. To determine if these modules positively impact performance, we set the basic ViT model as the Baseline. Our ablation experiments focus on the optimizer, triplet module, and loss function.

For the optimizer selection, we compared SGD[46]、Adam[47] and Adan[48]. Generally, Adam significantly improved model performance over SGD when it was introduced. However, our experiments revealed that SGD outperformed Adam in optimizing model parameters. The detailed experimental results are shown in Table II below:

TABLE II
IMPACT OF OPTIMIZERS ON MODEL TRAINING

| Setting | Rank-1 | Rank-5 | Rank-10 | mAP |
|---|---|---|---|---|
| PCDNet+Adan | 67.9 | 71.6 | 85.1 | 88.3 |
| PCDNet+Adam | 70.0 | 84.1 | 87.5 | 65.7 |
| **PCDNet+SGD** | **80.1** | **92.4** | **95.1** | **78.3** |

As shown in Table II, although the Adam optimizer can accelerate convergence, it falls short of SGD in both Rank-1 and mAP performance in this experiment. We attribute the superior performance of SGD to its ability to adjust the learning rate more stably. The learning rate decay strategy of SGD helps the model approach optimal solutions more precisely in later training stages, resulting in higher accuracy on the test set. Additionally, SGD does not rely on second-order derivative information, which may reduce optimization noise caused by inaccurate gradient estimation, making feature extraction more robust.

In this study, the SGD optimizer offers PCDNet greater stability and robustness during later training, contributing to better performance. This finding underscores that traditional optimizers can still be effective for specific tasks and architectures, highlighting the importance of a deep understanding and careful tuning of optimizers in model optimization.

To investigate the effect of the triplet module on model performance, we compared the ViT model without a triplet module (TA module only) with the TA+ViT model (PCDNet). This experiment includes small, base, and large versions with 12, 16, and 32 Attention Blocks, respectively.

TABLE III
IMPACT OF THE TRIPLET MODULE ON MODEL PERFORMANCE

| Setting | Rank-1 | Rank-5 | Rank-10 | mAP |
|---|---|---|---|---|
| ViT-small | 69.5 | 84.2 | 87.8 | 53.9 |
| ViT-base | 72.2 | 85.7 | 89.1 | 68.0 |
| ViT-large | 72.5 | 82.9 | 85.9 | 64.4 |
| PCDNet-small | 80.1 | 92.4 | 95.1 | 78.3 |
| PCDNet base - | 80.6 | 92.5 | 95.6 | 78.6 |
| **PCDNet large** - | **81.4** | **92.6** | **95.6** | **79.3** |

As shown in Table III, introducing the triplet module significantly enhances model performance across all ViT versions—small, base, and large—indicating its effectiveness in boosting feature extraction and recognition accuracy. Without the triplet module, model performance trends upward and then downward as model size increases (small to base to large), suggesting that larger models do not always yield better performance and that there may be an optimal model size. However, with the triplet module, all model versions show improved performance. Notably, adding the triplet module to the small version boosts Rank-1, Rank-5, Rank-10, and mAP by 10.5%, 8.1%, 7.2%, and 24.4%, respectively. This improvement is particularly evident in smaller models.

The large version of PCDNet with the triplet module performs best across all metrics, suggesting that, with the help of the triplet module, a larger model can better capture the dataset's features and improve performance.

These experimental results demonstrate the triplet module's substantial impact on improving ViT model performance, regardless of model size. With the triplet module, the large version outperforms the small and base versions slightly, possibly due to enhanced feature extraction capabilities in larger models. Future work can further explore the integration of the triplet module with other architectures and its performance on different datasets. Additionally, research could focus on optimizing the design of the triplet module to suit a broader range of applications. It should be noted that, while the triplet module shows positive effects in this study, these results are based on a specific dataset and task. The triplet module's effectiveness may vary with different data distributions and task requirements, necessitating validation on additional datasets.

### B. Label Smooth

In normal cross-entropy loss, we use one-hot encoding to represent target labels. For a given sample belonging to class $i$, the target label $y$ is a vector where the $i$-th element is 1, and all other elements are 0. The cross-entropy loss formula is:

$$L_{ce} = -\sum_{i=1}^{K} y_i \log(p_i) \quad (9)$$

where $K$ is the total number of classes, $y_i$ is the true label (usually one-hot encoded), and $p_i$ is the model's predicted probability. Label smoothing modifies the target label distribution, making it less definitively one-hot. Instead, it assigns a value slightly less than 1 to the correct class and a value greater than 0 but small to other classes. This approach aims to reduce overconfidence in the model's predictions. The label smoothing formula is:

$$y_{smoothed} = (1-\epsilon) \cdot y + \frac{\epsilon}{K} \quad (10)$$

where $\epsilon$ is a small constant, typically set to 0.1 or another value to control the smoothing level. $K$ represents the total number of classes, and $y$ is the original one-hot encoded target label. Substituting the smoothed label $y_{smoothed}$ into the cross-entropy loss formula, we obtain:

$$L_{LS} = -\sum_{i=1}^{K} y_{smoothed,i} \log(p_i) \quad (11)$$

The purpose of label smoothing is to reduce the model's overconfidence in the training data labels, thereby enhancing its generalization ability on new data. By introducing a small uncertainty $\epsilon$, the model is discouraged from being overly confident in any class, which improves its robustness when encountering uncertain or noisy data. In our experiments, using label smoothing $L_{LS}$ with $\epsilon = 0.1$ yields better model performance.

TABLE IV
IMPACT OF LABEL SMOOTHING ON MODEL PERFORMANCE

| Setting | Rank-1 | Rank-5 | Rank-10 | mAP |
|---|---|---|---|---|
| PCDNet (no label soomth) | 80.1 | 92.4 | 95.1 | 78.3 |
| **PCDNet (with label soomth)** | **82.7** | **92.2** | **94.5** | **79.0** |

In our experiments, we compared model performance under the same architecture, using traditional cross-entropy loss (without label smoothing) and the label-smoothed loss function. Results in Table IV indicate that label smoothing enhances the model's generalization, especially when handling uncertain or noisy data. Overall, label smoothing improves generalization on new data, making the model more robust in practical applications. While some metrics showed slight declines, the improvements in Rank-1 and mAP indicate that label smoothing effectively enhances model accuracy and robustness. These results support the use of label smoothing as a straightforward yet effective regularization strategy in model training.

### C. Model Interpretability

Due to the end-to-end black-box nature of deep learning, researchers often find it challenging to understand exactly what knowledge a model has learned. To address this, we employed GardCAM[49] and GradCAM++[50] tools to visualize PCDNet's attention areas, enabling insight into what the model is truly focusing on.

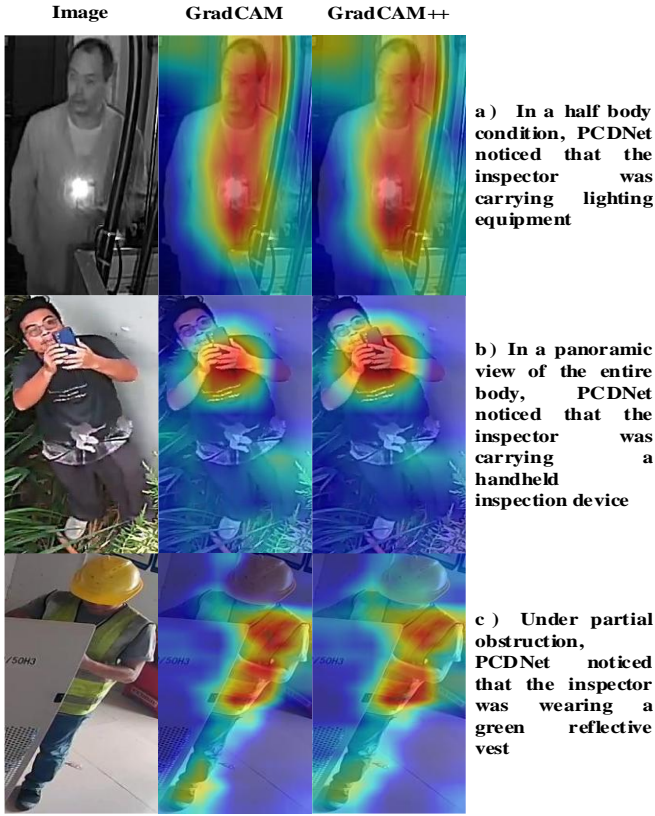

a) In a half body condition, PCDNet noticed that the inspector was carrying lighting equipment

b) In a panoramic view of the entire body, PCDNet noticed that the inspector was carrying a handheld inspection device

c) Under partial obstruction, PCDNet noticed that the inspector was wearing a green reflective vest

Fig. 6. presents visualizations of the first Transformer encoder in the penultimate layer of PCDNet using Grad-CAM and Grad-CAM++. In these heatmaps, red areas represent parts of the image that the model focuses on heavily, while blue areas represent parts with less focus. Grad-CAM utilizes average gradients to weight 2D activations, while Grad-CAM++ is similar but uses second-order gradients. The provided comparison examples reveal that PCDNet effectively focuses on meaningful component-level semantic information.

Examining Fig. 6 row by row, the first row shows a maintenance worker in a confined space with only the upper body exposed; however, the model successfully focuses on this worker's key body features. In the second row, a maintenance worker using a handheld device is inspecting equipment from below; PCDNet accurately identifies and focuses on the worker using the inspection device. In the third row, the maintenance worker is partially obscured, yet the model's focus remains on the worker rather than extending to the external equipment casing. This indicates that the model understands semantic information related to occlusion.

In summary, these three practical examples demonstrate that our model consistently attends to relevant component features, highlighting its potential for real-world deployment.

### D. Case Study

To enhance the practicality of PCD-ReID, we developed a program to observe PCDNet's stability in real-world applications. The following figure includes the Query image on the far left, followed by two rows of image data. The top row displays images from the Gallery that PCDNet matched with the Query image, arranged from left to right in decreasing similarity. The bottom row shows the known correct matches, enabling us to assess the model's accuracy. To provide richer information, each image is annotated with three indicators: similarity score, match status, and person ID. Gallery images are also framed, with red borders indicating successful matches and blue borders indicating mismatches.

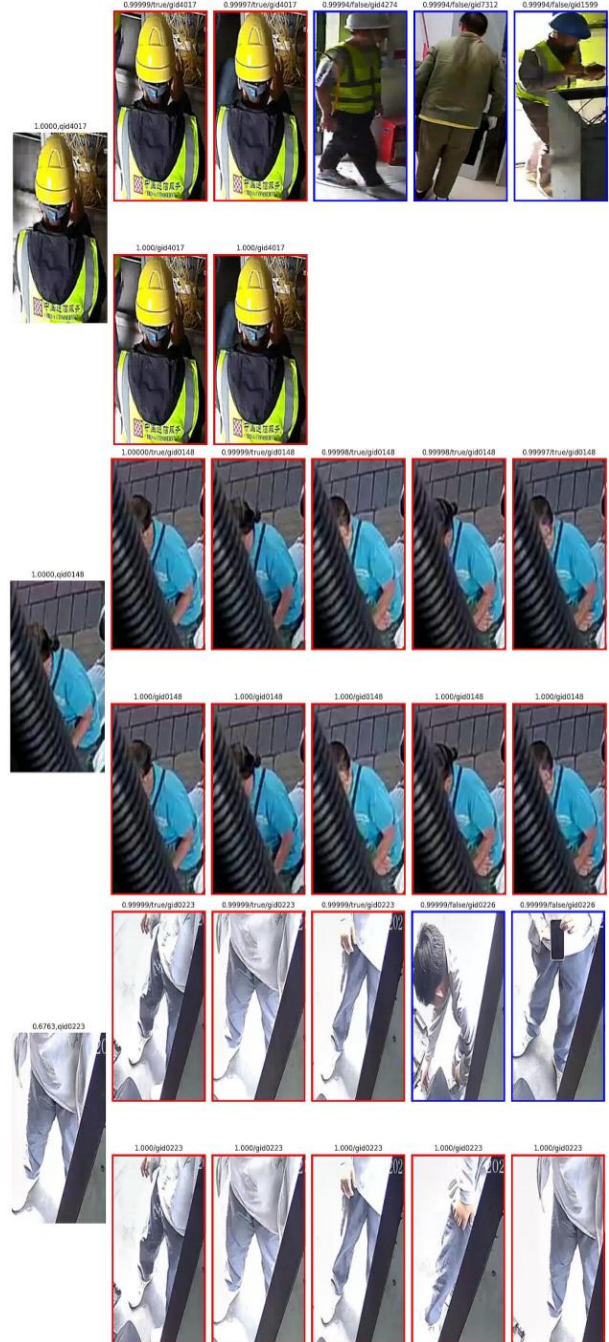

Fig. 7. Presents several cases demonstrating the model's performance in practical applications, each with a specific focus. The first case highlights PCD-ReID's robustness in matching individuals despite limited sample images. The second case illustrates stable matching even with occlusions, and the third case emphasizes the model's resilience when handling mislabeled data.

From these examples in Fig. 7, we observe that PCDNet exhibits stability in real-world use. When matching ID4017, although the individual's face and full body are not fully visible, PCDNet accurately identifies the person by focusing on helmet and upper-body vest features. Even when the Gallery does not contain this ID, the model still identifies two individuals with similar helmet features, indicating its ability to discern feature differences. In the case of ID0148, the model's focus on the distinctive blue top results in an excellent match. For ID0223, the model yields three correct matches and two incorrect ones by identifying ID0226; after verification, these IDs represent the same person. This outcome suggests that the model's ability to discern feature differences can help correct labeling errors.

In summary, our algorithm demonstrates exceptional stability in practical applications, showcasing its strong potential for real-world deployment.

## VI. CONCLUSION

This paper proposes a new method, PCD-ReID, for re-identifying base station inspection personnel, specifically addressing the challenges of pedestrian recognition in occluded base station environments. The main contributions are as follows:

**Transformer-Based PCD Network Design**: To tackle the occlusion problem, we propose a Pedestrian Component Differentiation Network (PCD-Net) that captures shared features such as helmets and uniforms, addressing the limitations of traditional pedestrian re-identification in occluded settings. Built on a Vision Transformer with a triplet attention module, this model effectively captures unoccluded local features, enhancing robustness in recognition.

**Dataset Construction and Multi-Loss Optimization**: We collected over 50,000 images of inspection personnel in real surveillance settings to create the MyTT2 dataset, with data augmentation to increase diversity. Additionally, we designed a joint loss function combining cross-entropy, triplet, Circle, and Cosface losses to optimize the model's discriminative power and stability.

**High Performance**: Testing in actual base station inspection scenarios shows that the PCD-ReID model achieves 79.0% mAP and 82.7% Rank-1, an improvement of 15.9% over traditional ResNet50, validating the model's effectiveness in base station personnel re-identification and its potential for real-world deployment.

Future efforts will explore feature extraction methods based on multimodal data and external semantic information to further improve robustness in complex occlusion scenarios. Additionally, given real-time requirements, model structure and inference speed will be optimized to facilitate the practical deployment of PCD-ReID for seamless check-in applications.